\documentclass[conference]{IEEEtran}
\IEEEoverridecommandlockouts
\usepackage{cite}
\usepackage{amsmath,amssymb,amsfonts}
\usepackage{algorithmic}
\usepackage{graphicx}
\usepackage{textcomp}
\usepackage{xcolor}
\usepackage{hyperref}
\def\BibTeX{{\rm B\kern-.05em{\sc i\kern-.025em b}\kern-.08em
    T\kern-.1667em\lower.7ex\hbox{E}\kern-.125emX}}
\begin{document}

\title{Global Context Modeling in YOLOv8 for Pediatric Wrist Fracture Detection\\
\thanks{This research is supported by National Science and Technology Council of Taiwan, under Grant Number: NSTC 112-2221-E-032-037-MY2.}}

\author{\IEEEauthorblockN{Rui-Yang~Ju}
\IEEEauthorblockA{
\textit{National Taiwan University}\\
Taipei City, Taiwan \\
jryjry1094791442@gmail.com}
\and
\IEEEauthorblockN{Chun-Tse~Chien}
\IEEEauthorblockA{
\textit{Tamkang University}\\
New Taipei City, Taiwan \\
popper0927@hotmail.com}
\and
\IEEEauthorblockN{Chia-Min~Lin}
\IEEEauthorblockA{
\textit{Tamkang University}\\
New Taipei City, Taiwan \\
813440038@o365.tku.edu.tw}
\and
\IEEEauthorblockN{Jen-Shiun~Chiang}
\IEEEauthorblockA{
\textit{Tamkang University}\\
New Taipei City, Taiwan \\
jsken.chiang@gmail.com}
}

\maketitle

\begin{abstract}
Children often suffer wrist injuries in daily life, while fracture injuring radiologists usually need to analyze and interpret X-ray images before surgical treatment by surgeons.
The development of deep learning has enabled neural network models to work as computer-assisted diagnosis (CAD) tools to help doctors and experts in diagnosis.
Since the YOLOv8 models have obtained the satisfactory success in object detection tasks, it has been applied to fracture detection.
The Global Context (GC) block effectively models the global context in a lightweight way, and incorporating it into YOLOv8 can greatly improve the model performance.
This paper proposes the YOLOv8+GC model for fracture detection, which is an improved version of the YOLOv8 model with the GC block.
Experimental results demonstrate that compared to the original YOLOv8 model, the proposed YOLOv8-GC model increases the mean average precision calculated at intersection over union threshold of 0.5 (mAP 50) from 63.58\% to 66.32\% on the GRAZPEDWRI-DX dataset, achieving the state-of-the-art (SOTA) level.
The implementation code for this work is available on GitHub at \url{https://github.com/RuiyangJu/YOLOv8_Global_Context_Fracture_Detection}.
\end{abstract}


\section{Introduction}
Wrist injuries are very common among children \cite{hedstrom2010epidemiology,randsborg2013fractures}.
Research \cite{meena2014fractures} has shown that most fractures occur 2 centimeters from the distal radius near the joint.
If they are not treated promptly and effectively, these injuries will result in wrist joint deformities, limited joint motion, and chronic pain \cite{bamford2010qualitative}.
In severe cases, an incorrect diagnosis would lead to lifelong complications and inconvenience \cite{kraus2010treatment}.

With the rapid development of deep learning, neural network models are increasingly utilized as computer-assisted diagnosis (CAD) tools to aid doctors in analyzing X-ray images.
Object detection models can accurately predict fractures and reduce the probability of misdiagnosis.
Although two-stage object detection models \cite{girshick2014rich,girshick2015fast,ren2015faster} obtain the excellent model performance, one-stage object detection models \cite{liu2016ssd,redmon2016you} offer faster inference times.
With the continued release of YOLO series models, YOLOv8 \cite{jocher2023yolo} and YOLOv9 \cite{wang2024yolov9} models enable efficient object detection on low-computing platforms \cite{ju2024resolution}.

\begin{figure}[t]
\centering
\includegraphics[width=\linewidth]{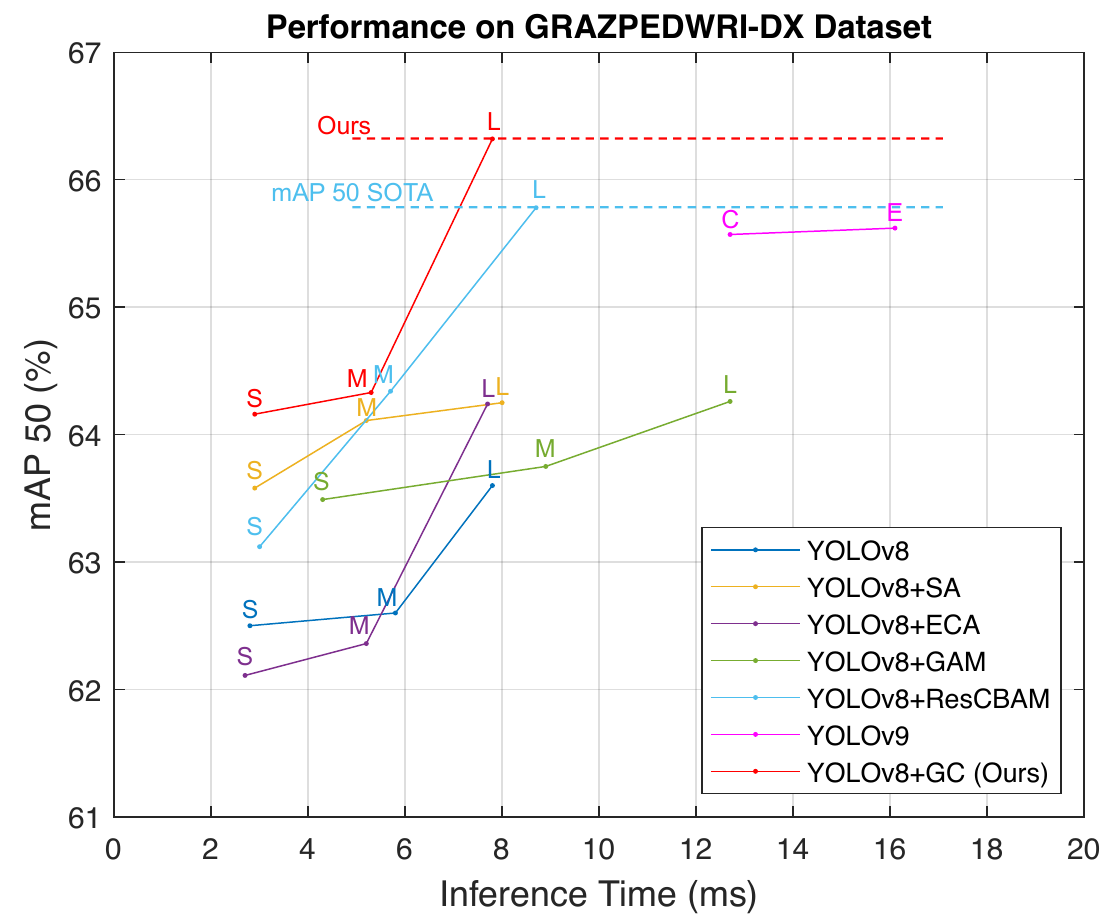}
\caption{Comparison of different models on the GRAZPEDWRI-DX dataset for fracture detection with the input image size of 1024. In terms of mAP 50, the proposed model YOLOv8+GC outperforms the YOLOv8+ResCBAM model, which is the SOTA model.}
\label{fig:intro}
\vspace{-4mm}
\end{figure}

Since the release of the GRAZPEDWRI-DX \cite{nagy2022pediatric} dataset by the Medical University of Graz, Ju \emph{et al.} \cite{ju2023fracture} have been the first to deploy the YOLOv8 model to detect fractures on this dataset.
This deployment helped surgeons to interpret fractures in X-ray images, reducing misdiagnosis and providing a better information base for the surgery.
Chien \emph{et al.} \cite{chien2024yolov8} developed the YOLOv8-AM models by integrating different attention modules \cite{woo2018cbam,wang2020eca,liu2021global,zhang2021sa} into YOLOv8 \cite{jocher2023yolo}, which improved the model performance in fracture detection.
Additionally, Chien \emph{et al.} \cite{chien2024yolov9} applied the YOLOv9 model to the this dataset, achieving the state-of-the-art (SOTA) model performance.
However, there remains a a great possibility for improving the model performance of the current SOTA model.
This work proposes the YOLOv8+GC model for fracture detection, which integrates the Global Context (GC) \cite{cao2020global} blocks to enhance the model performance, and the results are shown in Fig. \ref{fig:intro}.

\begin{figure*}[t]
\centering
\includegraphics[width=\linewidth]{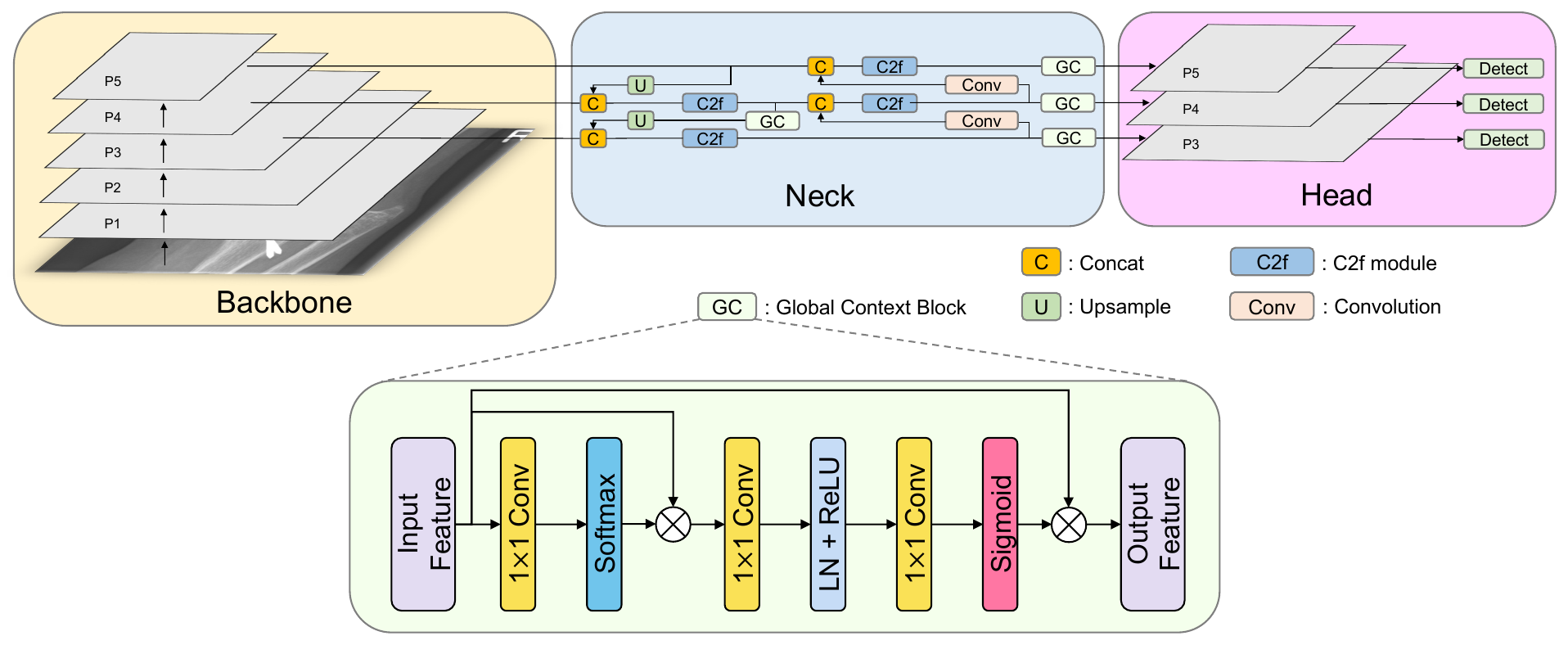}
\caption{The detailed illustration of the proposed YOLOv8+GC model architecture, including backbone, neck, and head. The GC block is designed to be added to the neck of the model architecture.}
\label{fig:arch}
\end{figure*}

The main contributions of this paper are as follows:
\begin{itemize}
\vspace{-1mm}
\item This work combines the YOLOv8 model with the GC block and proposes the YOLOv8+GC model, which improves the mAP 50 value by 4.3\% compared to the original YOLOv8 model.
\item The proposed model achieves the mAP 50 value of 66.32\% on the GRAZPEDWRI-DX dataset for fracture detection, reaching the SOTA level.
\item The proposed model can be utilized as CAD tools to assist doctors and experts in interpreting X-ray images of pediatric wrist injuries in real-world.
\end{itemize}

\section{Related Work}
Fracture detection is an important topic in medical image diagnostic tasks.
Publicly available datasets \cite{rajpurkar2017mura} for fracture detection in adults are limited, and even fewer datasets \cite{halabi2019rsna} are available for pediatrics.
Nagy \emph{et al} \cite{nagy2022pediatric} published a pediatric related dataset named the GRAZPEDWRI-DX dataset.
Hržić \emph{et al.} \cite{hrvzic2022fracture} demonstrated that the YOLOv4 \cite{bochkovskiy2020yolov4} model could enhance the accuracy of pediatric wrist injury diagnoses in X-ray images.
Ju \emph{et al.} \cite{ju2023fracture} developed an application employing the YOLOv8 model to assist doctors in interpreting X-ray images.
Chien \emph{et al.} \cite{chien2024yolov8} introduced four attention modules, including Convolutional Block Attention Module (CBAM) \cite{woo2018cbam}, Efficient Channel Attention (ECA) \cite{wang2020eca}, Global Attention Mechanism (GAM) \cite{liu2021global}, and Shuffle Attention (SA) \cite{zhang2021sa} to improve the model performance, where YOLOv8+ResCBAM achieving the SOTA model performance.
In addition, Chien \emph{et al.} \cite{chien2024yolov9} applied the YOLOv9 model \cite{wang2024yolov9} to this dataset, and also achieved the excellent model performance.

The GC block \cite{cao2020global} is a neural network module designed to improve the model's ability for understanding and processing global information.
By extracting global features from the whole image, this block improves the model's comprehension of the image overall contents.
It employs an adaptive mechanism to integrate local and global features to improve the model's recognition and presentation ability.
Furthermore, the GC block combines multi-scale features, improving the model's ability to process image contents at various scales and resolutions.
This capability enables the model to more effectively recognize and understand both the details and the global information of the image.
Based on this, this paper proposes the YOLOv8+GC model, which aims to further improve the model performance for fracture detection.

\begin{table*}[ht]
\centering
\caption{Experimental results of the ablation study of the YOLOv8 models based on the global context block for fracture detection}
\label{tab:ablation}
\setlength{\tabcolsep}{3.8mm}{
\begin{tabular}{lcccccccc}
\hline \noalign{\smallskip}
\textbf{Method} & \textbf{Model Size} & \textbf{Input Size} & \textbf{Params} & \textbf{FLOPs} & \textbf{F1-Score} & \textbf{mAP $_{50}\rm ^{val}$} & \textbf{mAP $_{50-95}\rm ^{val}$} & \textbf{Inference Time} \\ \noalign{\smallskip} \hline \noalign{\smallskip} \hline \noalign{\smallskip}
Baseline & Small & 1024 & 11.13M & 28.5G & 0.60 & 62.5\% & 39.9\% & 2.8ms \\
Baseline & Medium & 1024 & 25.84M & 78.7G & 0.61 & 62.6\% & 40.1\% & 5.2ms \\
Baseline & Large & 1024 & 43.61M & 164.9G & 0.62 & 63.6\% & 40.4\% & 7.7ms \\ \noalign{\smallskip} \hline \noalign{\smallskip}
+GC Block & Small & 1024 & 11.24M & 28.7G & 0.64 & 64.2\% & 40.6\% & 2.8ms \\
+GC Block & Medium & 1024 & 26.03M & 79.2G & 0.65 & 64.3\% & 41.0\% & 5.2ms \\
+GC Block & Large & 1024 & 43.85M & 165.6G & 0.66 & 66.3\% & 42.9\% & 7.9ms \\ \noalign{\smallskip} \hline \noalign{\smallskip}
\end{tabular}}
\end{table*}

\begin{table*}[t]
\centering
\caption{Experimental results of fracture detection on the grazpedwri-dx dataset for the proposed model and different models}
\label{table_comparison}
\setlength{\tabcolsep}{3.2mm}{
\begin{tabular}{lcccccccc}
\hline \noalign{\smallskip}
\textbf{Method} & \textbf{Model Size} & \textbf{Input Size} & \textbf{Params} & \textbf{FLOPs} & \textbf{F1-Score} & \textbf{mAP $_{50}\rm ^{val}$} & \textbf{mAP $_{50-95}\rm ^{val}$} & \textbf{Inference Time} \\ \noalign{\smallskip} \hline \noalign{\smallskip} \hline \noalign{\smallskip}
YOLOv8 & Large & 1024 & 43.61M & 164.9G & 0.62 & 63.58\% & 40.40\% & 7.7ms \\
YOLOv8+SA & Large & 1024 & 43.64M & 165.4G & 0.63 & 64.25\% & 41.64\% & 8.0ms \\
YOLOv8+ECA & Large & 1024 & 43.64M & 165.5G & 0.65 & 64.24\% & 41.94\% & 7.7ms \\
YOLOv8+GAM & Large & 1024 & 49.29M & 183.5G & 0.65 & 64.26\% & 41.00\% & 12.7ms \\
YOLOv8+ResGAM & Large & 1024 & 49.29M & 183.5G & 0.64 & 64.98\% & 41.75\% & 18.1ms \\
YOLOv8+ResCBAM & Large & 1024 & 53.87M & 196.2G & 0.64 & {\color{blue}65.78\%} & 42.16\% & 8.7ms \\
YOLOv9 & Compact & 1024 & 51.02M & 239.0G & 0.66 & 65.57\% & 43.70\% & 12.7ms \\
YOLOv9 & Extended & 1024 & 69.42M & 244.9G & 0.66 & 65.62\% & {\color{red}43.73\%} & 16.1ms \\
YOLOv8+GC (Ours) & Large & 1024 & 43.85M & 165.6G & 0.66 & {\color{red}66.32\%} & {\color{blue}42.85\%} & 7.9ms \\ \noalign{\smallskip} \hline \noalign{\smallskip}
\multicolumn{9}{l}{Best and 2nd best performance are in {\color{red}red} and {\color{blue}blue} colors, respectively.}
\end{tabular}}
\end{table*}

\section{Method}
\subsection{Network Architecture}
Chien \emph{et al.} \cite{chien2024yolov8} demonstrated that incorporating various modules into the YOLOv8 model significantly improve the model performance.
This paper introduces the GC block into the Neck part of the YOLOv8 model architecture to improve global feature extraction and capture comprehensive image information.
Specifically, as illustrated in Fig. \ref{fig:arch}, this work adds one GC block after each of the four C2f modules (cross stage partial bottlenecks with two convolutions).

\subsection{Global Context Block}
The GC block \cite{cao2020global} is highly suitable for fracture detection tasks due to its flexible properties and ease of integration into different neural network architectures, including the YOLOv8 \cite{jocher2023yolo} model.
This block combines the advantages of the Simplified Non-Local (SNL) \cite{cao2020global} block, which effectively models long-range dependencies, and the Squeeze-Excitation (SE) \cite{hu2018squeeze} block, known for its computational efficiency.
As shown in Fig. \ref{fig:arch}, the GC block follows the Non-Local (NL) \cite{wang2018non} block and aggregates the global context across all locations, enabling the capture of long-range dependencies.
Furthermore, the GC block is lightweight and can be applied to multiple layers, enhancing the model's ability to capture long-range dependencies with only a slight increase in computational cost.
Therefore, this work integrates the GC block into the YOLOv8 model architecture for fracture detection.

\section{Experiments}
\subsection{Dataset and Evaluation Metrics}
GRAZPEDWRI-DX \cite{nagy2022pediatric} dataset is a publicly available dataset of pediatric wrist trauma X-ray images provided by the Medical University of Graz.
Collected by radiologists from the University Hospital Graz between 2008 and 2018, this dataset comprises 20,327 X-ray images.
Nagy \emph{et al.} \cite{nagy2022pediatric} highlight that prior to the release of this dataset, there are few publicly available relevant pediatric datasets. 

Fracture detection in pediatric wrist X-ray images belongs to the object detection task.
Six evaluation metrics used in object detection are applicable, including the model parameters (Params), floating point operations (FLOPs), F1-score, the mean average precision calculated at intersection over union threshold of 0.5 (mAP 50), the mean average precision calculated at intersection over union thresholds from 0.5 to 0.95 (mAP 50-95), and the inference time.
Therefore, this work compares the performance of different models using these evaluation metrics.

\subsection{Data Preparation}
To ensure fairness in model performance comparison, the same dataset is used for all model training.
This work randomly divide the GRAZPEDWRI-DX \cite{nagy2022pediatric} dataset into training, validation, and test sets in the ratio of 70\%, 20\%, and 10\%, respectively.
Specifically, the training set contains 14,204 images (69.88\%); the validation set contains 4,094 images (20.14\%), and the test set contains 2,029 images (9.98\%).

Furthermore, data augmentation is performed on the training set before model training for all models.
Specifically, this work adjusts the contrast and brightness of the images using the addWeighted function in the Open Source Computer Vision Library (OpenCV), and expanding the original training set from 14,204 images to 28,408 images.

\subsection{Training}
To ensure a fair comparison of the performance of different models on the GRAZPEDWRI-DX \cite{nagy2022pediatric} dataset for fracture detection, the same training environment and hardware (NVIDIA GeForce RTX 3090 GPUs) are used for model training and evaluation.

This work employs Python 3.9 and the PyTorch framework for model training.
For the hyperparameters of model training, this work sets the batch size to 16 and the epochs of 100.
In addition, the models are trained using the SGD optimizer, with the weight decay of 0.0005, momentum of 0.937, and the initial learning rate of 0.01.

\begin{figure*}[t]
\centering
\includegraphics[width=\linewidth]{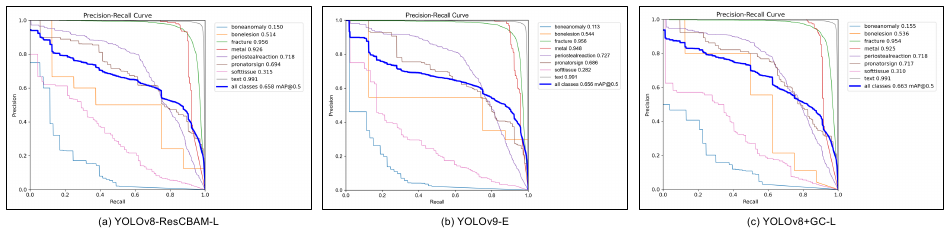}
\caption{Visualization of the accuracy of predicting each class using YOLOv8+ResCBAM-L model, YOLOv9-E model, and the proposed model YOLOv8+GC-L on the GRAZPEDWRI-DX dataset with the input image size of 1024.}
\label{fig:result}
\end{figure*}

\begin{figure*}[t]
\centering
\includegraphics[width=\linewidth]{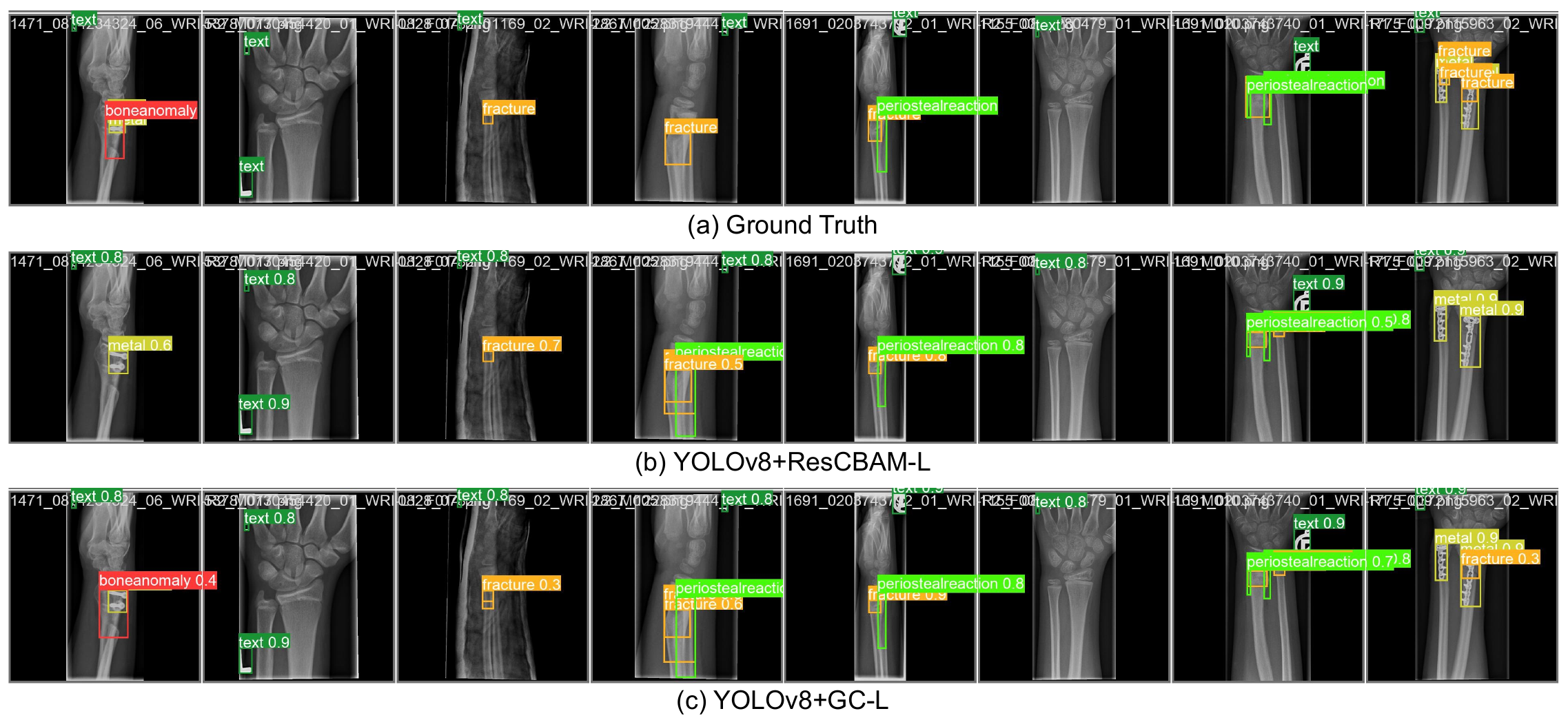}
\caption{Examples of prediction results of the YOLOv8+ResCBAM-L model and the proposed model YOLOv8+GC-L model for fracture detection with the input image size of 1024.}
\label{fig:predict}
\end{figure*}

\subsection{Ablation Study}
It is well known that larger size models generally obtain the better model performance, including the improvements in mAP 50, mAP 50-95, along with longer inference times.
As shown in Table \ref{tab:ablation}, when the input image size is 1024, the proposed 
model containing GC blocks outperforms the original YOLOv8 model in all model sizes, with lesser increases in inference time and model size.
Specifically, for the large model size, the mAP 50 of the YOLOv8+GC model improves from 63.6\% (YOLOv8) to 66.32\%, while the inference time increases by only 0.2 ms per image.
Additionally, the model's Params and FLOPs increase slightly, only from 43.61M to 43.85M and 164.9G to 165.6G, respectively.

\subsection{Experimental Results}
For comparative analysis with other models, this work sets the input image size to 1024 and adjusts the size of different YOLOv8 models to large (L) and YOLOv9 models to compact (C) and extended (E), respectively.
The experimental results in Table \ref{table_comparison} demonstrate that the proposed model YOLOv8+GC achieves the same F1-score as the YOLOv9 model, which is 0.66, reaching the SOTA level.
Additionally, the proposed model obtains the mAP 50 value of 66.32\%, surpassing 65.78\% of the previous SOTA model YOLOv8+ResCBAM.
Furthermore, the proposed model achieves the second highest mAP 50-95 value.
Based on this, the inference time of the proposed model YOLOv8+GC is significantly shorter at 7.9 ms, compared to 8.7 ms of YOLOv8+ResCBAM and 16.1 ms of YOLOv9.
For the model's Params and FLOPs, the proposed model YOLOv8+GC is also more efficient than the previous SOTA models.
These experimental results demonstrate the positive effect of the model performance with the addition of the GC block to the YOLOv8 model architecture.

Fig. \ref{fig:result} presents the Precision-Recall Curve (PRC) of the proposed model and the previous SOTA models for each class.
According to Fig. \ref{fig:result}, all models have good ability to correctly detect the ``fracture'', ``metal'', and ``text'' classes, with the accuracy all above 90\%.
However, the poor ability of these models to detect the ``bone anomaly'' class seriously affects the mAP 50 values of the models.
The proposed model detects the ``bone anomaly'' class with the accuracy of 15.5\%, compared to 15.0\% of YOLOv8+ResCBAM, and 11.3\% of YOLOv9.

In a real-world diagnostic scenario, this paper evaluates the impact of the GC block on the performance of the YOLOv8 model for fracture detection.
Eight X-ray images are randomly selected, and the prediction results of the YOLOv8+ResCBAM and YOLOv8+GC models are shown in Fig. \ref{fig:predict}. 
Compared to the previous SOTA model YOLOv8+ResCBAM, the proposed model can detect the wrist fracture more accurately in complex injury situations.
Specifically, in the rightmost X-ray image of Fig. \ref{fig:predict}, the proposed model YOLOv8+GC can correctly detect the fracture, while the YOLOv8+ResCBAM model cannot.

\section{Discussion}
Compared to the original YOLOv8 model, although the mAP 50 value of the proposed model improved from 63.58\% to 66.32\%, the mAP 50 value still did not reach 70\%.
This issue is primarily due to the poor ability of the model in predicting the ``bone anomaly'' and ``soft tissue'' classes.
The dataset suffers from the class imbalance, for example, compared to 23,722 and 18,090 labels of the ``text'' and ``fracture'' classes, but there are only 276 and 464 labels of the ``bone anomaly'' and ``soft tissue'' classes.
The performance of the YOLO series models mainly depends on the quality and diversity of the training set.
The limited training samples for these underrepresented classes result in the poor prediction ability of the model.
Therefore, to further improve the model performance, obtaining more data for the ``bone anomaly'' and ``soft tissue'' classes is important.

\section{Conclusion}
Applying the YOLOv8, YOLOv8-AM, and YOLOv9 models to pediatric wrist fracture detection has resulted in the model performance breakthroughs on the GRAZPEDWRI-DX dataset.
The current SOTA models, YOLOv8+ResCBAM and YOLOv9, achieve the mAP 50 values of 65.78\% and 65.62\%, and the mAP 50-95 values of 42.16\% and 43.73\%, respectively.
However, these models are significantly larger than the original YOLOv8 model in terms of model's Params and FLOPs.
For example, the original YOLOv8-L model has only 43.61M parameters, while YOLOv8+ResCBAM-L and YOLOv9-E models have 53.87M and 69.42M parameters, respectively, which is not satisfactory.
To address this, this paper proposes the YOLOv8+GC model, which improves the model performance efficiently in a lightweight way.
The proposed model YOLOv8+GC has only 43.85M parameters and achieves the mAP 50 value of 66.32\%, which exceeds the current SOTA level.

\bibliographystyle{IEEEtran}
\bibliography{main}
\end{document}